\documentclass[letterpaper, 10 pt, conference]{ieeeconf}
\IEEEoverridecommandlockouts  
\usepackage{amsmath,amsfonts}
\usepackage{algorithmic}
\usepackage{algorithm}
\usepackage{array}
\usepackage[caption=false,font=normalsize,labelfont=sf,textfont=sf]{subfig}
\usepackage{yfonts}
\usepackage{textcomp}
\usepackage{stfloats}
\usepackage{url}
\usepackage{verbatim}
\usepackage{graphicx}
\usepackage{cite}
\usepackage{arydshln} 

\usepackage{mathtools}
\usepackage{upgreek}
\usepackage{stmaryrd}

\usepackage{tikz}
\usetikzlibrary{3d, arrows.meta}
\usetikzlibrary{calc}

\usetikzlibrary{arrows.meta, positioning, shapes.geometric}
 \usetikzlibrary{shapes.geometric, arrows} \tikzstyle{startstop} = [rectangle, rounded corners, minimum width=2cm, minimum height=1cm,text centered, draw=black, fill=pink!45!gray!20] \tikzstyle{process} = [rectangle, minimum width=2cm, minimum height=1cm, text centered, draw=black, fill=pink!45!gray!20] \tikzstyle{decision} = [diamond, minimum width=1cm, minimum height=1cm, text centered, draw=black, fill=pink!45!gray!20] \tikzstyle{arrow} = [thick,->,>=stealth]

\newcommand{\vect}[1]{\boldsymbol{\mathbf{#1}}}

\newcommand{\Bvect}[1]{\Bar{\boldsymbol{\mathbf{#1}}}}

\begin{document}

\title{\LARGE \bf
Correlation-Aware Dual-View Pose and Velocity Estimation for Dynamic Robotic Manipulation
}
\author{Mahboubeh Zarei, Robin Chhabra, and Farrokh Janabi-Sharifi
\thanks{Authors are with the Department of Mechanical, Industrial and Mechatronics Engineering, Toronto Metropolitan University, Toronto, Canada
        {\tt\small robin.chhabra@torontomu.ca}}%
\thanks{
        {\tt\small fsharifi@torontomu.ca }}%
\thanks{
        {\tt\small zarei@torontomu.ca }}%
}


\maketitle
\thispagestyle{empty}
\pagestyle{empty}

\begin{abstract}
Accurate pose and velocity estimation is essential for effective spatial task planning in robotic manipulators. While centralized sensor fusion has traditionally been used to improve pose estimation accuracy, this paper presents a novel decentralized fusion approach to estimate both pose and velocity. We use dual-view measurements from an eye-in-hand and an eye-to-hand vision sensor configuration mounted on a manipulator to track a target object whose motion is modeled as random walk (stochastic acceleration model). The robot runs two independent adaptive extended Kalman filters formulated on a matrix Lie group, developed as part of this work. These filters predict poses and velocities on the manifold $\mathbb{SE}(3) \times \mathbb{R}^3 \times \mathbb{R}^3$ and update the state on the manifold $\mathbb{SE}(3)$. The final fused state comprising the fused pose and  velocities of the target is obtained using a correlation-aware fusion rule on Lie groups. The proposed method is evaluated on a UFactory xArm 850 equipped with Intel RealSense cameras, tracking a moving target. Experimental results validate the effectiveness and robustness of the proposed decentralized dual-view estimation framework, showing consistent improvements over state-of-the-art methods.

\end{abstract}


\section{Introduction}
 Position-Based Visual Servoing (PBVS) for robotic manipulators refers to the problem of  controlling a robot's end-effector by minimizing the error between its current and desired pose (i.e., position and orientation) relative to a target object using vision-based data obtained from a set of cameras~\cite{sanderson1980image}. Compared to Image-Based Visual Servoing (IBVS) that regulates the error signal in image space, PBVS offers simpler 3-D control laws, greater robustness, provided that calibration and pose estimation are accurate, and easier integration with spatial tasks like path planning and collision avoidance~\cite{hager1996tutorial}. A comprehensive comparison between PBVS and IBVS is presented in~\cite{janabi2010comparison}, with a focus on system stability, robustness, sensitivity, and dynamic performance in both 3-D and image spaces.  The first step in PBVS is to accurately estimate the pose of the target object relative to a pre-defined reference coordinate frame. 
The classical vision-based approach to 6-DoF pose estimation typically involves extracting local features from the input image, matching them to those of a known 3-D model, and estimating the pose using algorithms such as Perspective-n-Point (PnP), RANSAC-based alignment, or Hough voting. While effective in controlled scenarios, these methods often struggle in real-world environments due to their sensitivity to occlusion, background clutter, lighting variations, and their reliance on carefully designed hand-crafted features. In recent years, significant progress has been made with deep learning and data-driven approaches, which address many of these limitations by learning robust features and pose representations directly from data~\cite{labbe2020cosypose,wen2020se}.

An important consideration to accurately obtain the pose  is the configuration of the vision sensors in the robot work space. In the context of robotic manipulators, cameras are typically arranged in one of three primary configurations: (i) eye-in-hand, where the cameras are rigidly mounted on the robot arm or end-effector and move with it~\cite{janabi1998automatic,janabi2010kalman,liang2024adaptive,he2022deep,ribeiro2023second}; (ii) eye-to-hand, where the cameras are fixed in the environment or on the robot base, allowing them to observe the target object and part or all of the manipulator from a stationary viewpoint~\cite{bauml2010kinematically,daravina2024visual,maniatis2017best}; and (iii) hybrid configurations, which combine both eye-in-hand and eye-to-hand setups to leverage the complementary strengths of each~\cite{flandin2000eye,kermorgant2011multi,dune2007one}. These systems can dynamically switch between camera sources or fuse information from both, improving overall accuracy, robustness, flexibility, and coverage. Several hybrid approaches have been proposed to exploit the advantages of multiple visual perspectives for real-time pose estimation and control. In~\cite{assa2014virtual}, a virtual visual servoing data fusion technique is introduced to estimate object pose using a set of eye-to-hand and eye-in-hand cameras. Similarly,\cite{lippiello2007position} presents a PBVS method that employs a hybrid eye-in-hand/eye-to-hand configuration, using an extended Kalman filter and an occlusion-aware feature selection algorithm to ensure accurate and efficient pose estimation. A more recent hybrid approach in\cite{cuevas2018hybrid} addresses the challenges of tracking a dynamically moving target by combining four eye-to-hand RGB-D sensors with an eye-in-hand stereo camera. A supervisory module dynamically switches between the two configurations for the pose estimation task based on the robot–target distance and visibility conditions. Although previous methods have explored sensor-level data fusion and switching mechanisms for pose estimation between eye-in-hand and eye-to-hand configurations, the challenge of consistently fusing pose estimates from multiple cameras in dynamic robotic manipulation remains unaddressed.   

To address these shortcomings, we propose a hybrid eye-to-hand/eye-in-hand robotic system in which each camera independently estimates the pose of a moving target using an adaptive Extended Kalman Filter (EKF) on Lie groups. The resulting estimates, expressed in the matrix Lie group $\mathbb{SE}(3)\times \mathbb{R}^3\times \mathbb{R}^3$, are then fused using a novel pose/velocity-level fusion strategy specifically designed to handle stochastic poses~\cite{zarei2024consistent}.  Therefore, the main contributions of this paper are as follows:
\begin{itemize}
    \item We develop a correlation-aware decentralized  fusion framework for dual-view target state estimation.    
    \item The estimation is propagated on the novel matrix Lie group $\mathbb{SE}(3)\times \mathbb{R}^3\times \mathbb{R}^3$, enabling joint inference of target pose and velocity using pose measurements on $\mathbb{SE}(3)$, as an alternative to existing approaches that rely solely on sensor-level fusion~\cite{assa2014virtual,lippiello2007position} or switching strategies~\cite{cuevas2018hybrid}.
    
    \item Building on the filter proposed in~\cite{bourmaud2013discrete}, we develop a new adaptive Lie group-based EKF to enhance robustness against measurement noise and unexpected uncertainties in the target object and robot's motion.

\end{itemize}
 

\color{blue}









\color{black}






\color{black}

\section{Preliminaries}\label{sec-pre}

\subsection{Matrix Lie groups in Robotics}

Here, the pose is described by a  matrix $\vect{\mathcal{T}}\in\mathbb{R}^{4\times4}$ that belongs to the 
 Special Euclidean Group~\cite{chirikjian2011stochastic}:
\begin{align}
    \mathbb{SE}(3)\coloneqq\begin{Bmatrix}\!\vect{\mathcal{T}}\!=\! \left[\begin{smallmatrix}\vect{\mathcal{R}}&\vect{p}\\\vect{0}_{3\times 1}^\top
&1
\end{smallmatrix}\right]|\vect{\mathcal{R}}\in\mathbb{SO}(3), \vect{p}\in\mathbb{R}^3 \end{Bmatrix},
\end{align}
where $\vect{p}$ is a translation vector with respect to some reference frame, and $\vect{\mathcal{R}}$ is a rotation matrix that belongs to the Special Orthogonal group $    \mathbb{SO}(3)\coloneqq\begin{Bmatrix}
         \vect{\mathcal{R}}\in\mathbb{R}^{3\times3}|\vect{\mathcal{R}}\vect{\mathcal{R}}^\top=\vect{I}_{3},~\det{(\vect{\mathcal{R}})}=1
     \end{Bmatrix}$. Here, $\vect{I}_n$ and $\vect{0}_n$ denote an $n$-dimensional identity matrix and zero matrix, respectively. 
The tangent space at the identity element $\vect{I}_{3}$  along with a bracket operator (commutator) forms the Lie algebra $\mathfrak{se}(3)$: 
\begin{align*}
    \mathfrak{se}(3)\!\coloneqq\!\begin{Bmatrix}\!
       \left[\vect{\zeta}\right]_\wedge\!=\!\left[\!\begin{smallmatrix}
           \left[\vect{\varphi}\right]_{\times}&\vect{\rho}\\\vect{0}_{3\times 1}^\top
&0        \end{smallmatrix}\!\right]\footnotesize\!\in\!\mathbb{R}^{4\times4}\!\bigg|  \vect{\varphi}, \vect{\rho}\in \mathbb{R}^3  
    \!\end{Bmatrix},
\end{align*}
where the operator $\left[.\right]_\wedge$ indicates the isomorphism between  $\mathbb{R}^{6}$ and $\mathfrak{se}_(3)$, with the inverse operator being denoted by $\left[.\right]_\vee$. That is, $\vect{\zeta}=\left[\begin{smallmatrix}
     \vect{\varphi}\\ \vect{\rho}  
\end{smallmatrix}\right]\in \mathbb{R}^{6}$. %
 Moreover, the operator $\left[.\right]_\times$ generates the skew-symmetric matrix of the vector $\vect{\varphi}$
 and $\left[.\right]_\otimes$ does the inverse.

 The exponential map $\exp: \mathfrak{se}(3)\rightarrow \mathbb{SE}(3)$ maps a Lie algebra element $\left[\vect{\zeta}\right]_\wedge\in\mathfrak{se}(3)$ to an element $\vect{\mathcal{T}}\in \mathbb{SE}(3)$ whose inverse is called the logarithm mapping $\log: \mathbb{SE}(3)\rightarrow \mathfrak{se}(3)$, with explicit forms: 
\begin{align}\nonumber
    \vect{\mathcal{T}}&\!=\!\exp(\left[\vect{\zeta}\right]_\wedge)\!=\!\left[\begin{smallmatrix}\exp(\left[\vect{\varphi}\right]_{\times})&\vect{V}(\vect{\varphi})\vect{\rho}\\\vect{0}_{3\times1}^\top&1\end{smallmatrix}\right],\\\nonumber
    \vect{\zeta}&=\left[\log(\vect{\mathcal{T}})\right]_\vee=\begin{bmatrix}
        \left[\log(\vect{\mathcal{R}})\right]_{\otimes}\\ \vect{V}^{-1}(\left[\log(\vect{\mathcal{R}})\right]_{\otimes})\vect{p}
    \end{bmatrix}.
\end{align}

Here, for non-zero $\phi\coloneqq  \lVert \vect{\varphi} \rVert\neq 0$, it is derived~\cite{chirikjian2011stochastic}:
\begin{align}
\vect{V}(\vect{\varphi})&=\vect{I}_{3}+\frac{1-\cos\phi}{\phi^2}\left[\vect{\varphi}\right]_{\times}+\frac{\phi-\sin\phi}{\phi^3}\left[\vect{\varphi}\right]^2_{\times},\\
  \vect{\mathcal{R}}\!=\!\exp&(\left[\vect{\varphi}\right]_{\times})\!=\!\vect{I}_{3}\!+\!\frac{\sin\phi}{\phi}\left[\vect{\varphi}\right]_{\times}\!+\!\frac{1-\cos\phi}{\phi^2}\left[\vect{\varphi}\right]_{\times}^2,
\end{align}
 and when  $\phi=0$, we have $ \vect{\mathcal{R}}=\vect{V}(\vect{\varphi})=\vect{I}_{3}$. Moreover, for  $\psi=\cos^{-1}(\frac{tr(\vect{\mathcal{R}})-1}{2})$, we have $\vect{\varphi}=\left[\log(\vect{\mathcal{R}})\right]_{\otimes}=\left[\frac{\psi}{2\sin\psi}(\vect{\mathcal{R}}-\vect{\mathcal{R}}^\top)\right]_\otimes.$
 

For an element $\vect{\mathcal{T}}\in \mathbb{SE}(3)$, the Adjoint $\vect{Ad}_{\vect{\mathcal{T}}}: \mathfrak{se}(3)\rightarrow \mathfrak{se}(3) $ is defined as a mapping that represents the conjugate action of $\mathbb{SE}(3)$ on its Lie algebra, i.e., $\vect{Ad}_{\vect{\mathcal{T}}}(\vect{\zeta})\coloneqq [\vect{\mathcal{T}}\left[\vect{\zeta}\right]_\wedge\vect{\mathcal{T}}^{-1}]_\vee$. Thus, the Adjoint representation of $\mathbb{SE}(3)$ takes the matrix form~\cite{murray2017mathematical,chirikjian2011stochastic}:
\begin{align}
 \small  \vect{Ad}_{\vect{\mathcal{T}}}\!=\!\begin{bmatrix}
        \vect{\mathcal{R}}& \vect{0}_{3}\\\left[\vect{p}\right]_{\times}\vect{\mathcal{R}}&\vect{\mathcal{R}}
        \end{bmatrix}\!\in\! \mathbb{R}^{6\times6}.
\end{align}

we can also define the adjoint representation of the Lie algebra $\mathfrak{se}(3)$ in the matrix form as follows:
\begin{align}
    \vect{ad}_{\vect{\zeta}}\!=\!\footnotesize\!\begin{bmatrix}
     \left[ \vect{\varphi}\right]_{\times}& \vect{0}_{3}\\\left[\vect{\rho}_1\right]_{\times}&\left[\vect{\varphi}\right]_{\times}    \end{bmatrix}\!\in\!\mathbb{R}^{6\times 6}.
\end{align}

The right Jacobian of  $\mathbb{SE}(3)$ for non-zero rotation angles is~\cite{barfoot2017state,chirikjian2011stochastic}:
\begin{align}\nonumber
&\mathcal{J}_r(\vect{\zeta})=\vect{I}_{6}-\frac{4-\phi \sin(\phi)-4\cos(\phi)}{2\phi^2}\vect{ad}_{\vect{\zeta}}+\\&\frac{4\phi-5\sin(\phi)+\phi\cos(\phi)}{2\phi^3}\vect{ad}_{\vect{\zeta}}^2-\nonumber\frac{2-\phi\sin(\phi)-2\cos(\phi)}{2\phi^4}\vect{ad}_{\vect{\zeta}}^3\\&+\frac{2\phi-3\sin(\phi)+\phi\cos(\phi)}{2\phi^5}\vect{ad}_{\vect{\zeta}}^4.
\end{align}
When the rotation angle is zero,  $\mathcal{J}_r(\vect{\zeta})=\vect{I}_{6}+\frac{1}{2}\vect{ad}_{\vect{\zeta}}$. 

\subsection{ Fusion on Matrix Lie Groups }\label{sec: configuration fusion}
The fusion problem aims to find an optimum consistent fused member $ \vect{\mathcal{X}}^\star_\text{fus}=\Bvect{\mathcal{X}}^\star_\text{fus}\exp([\vect{\zeta}_\text{fus}]_\wedge)\in\mathcal{G_X}$ with $\vect{\zeta}_\text{fus}\sim\mathcal{N}(\vect{0}_{\mathfrak{m}\times1},\vect{P}_\text{fus})$  from a given set of $n$ consistent and correlated stochastic members  $\vect{\mathcal{X}}_i=\Bvect{\mathcal{X}}_i\exp([\vect{\zeta}_i]_\wedge)\in\mathcal{G_X}$  with $\vect{\zeta}_i\sim\mathcal{N}(\vect{0}_{\mathfrak{m}\times 1},\vect{P}_{ii})$.  
The optimal fused member can be obtained from the following minimization problem~\cite{zarei2024consistent}:
\begin{align}\label{eq: opt problem 1}
 \vect{\mathcal{X}}^\star_\text{fus}=\arg \min_{\vect{\mathcal{X}}_\text{fus}}\begin{bmatrix}
 \vect{\varepsilon}_1\\
 \vdots\\
  \vect{\varepsilon}_n
\end{bmatrix}^\top\begin{bmatrix}
        \vect{P}_{11}&\cdots&\vect{P}_{1n}\\\vdots&\ddots&\vdots\\\vect{P}_{1n}^\top&\cdots&\vect{P}_{nn}\end{bmatrix}^{-1}\begin{bmatrix}
 \vect{\varepsilon}_1\\
 \vdots\\
  \vect{\varepsilon}_n
\end{bmatrix}, \end{align}
where the error between the $i^{th}$ known member and the fused member is 
$\vect{\varepsilon}_i\!=\!\left[\log(\Bvect{\mathcal{X}}_i^{-1}\vect{\mathcal{X}}_\text{fus})\right]_\vee\in\mathbb{R}^{\mathfrak{m}}$.
The closed-form solution to this minimization problem is presented in~\cite{zarei2024consistent}. For brevity, we only present the final derivation, as follows:
\begin{align}\label{eq: fused track final}
\Bvect{\mathcal{X}}_\text{fus}&=\Bvect{\mathcal{X}}_s\exp(\left[\Bvect{\zeta}_\text{fus}^\star\right]_\wedge),
\\ \label{eq: fused cov final}\vect{P}_\text{fus}&=\mathcal{J}_r(\Bvect{\zeta}_\text{fus}^\star)\vect{P}_{\vect{\zeta}\vect{\zeta}}^\star\mathcal{J}_r^\top(\Bvect{\zeta}_\text{fus}^\star).
\end{align}
Here, $\Bvect{\mathcal{X}}_s$ is an arbitrary reference member and
\begin{align}\nonumber
  \Bvect{\zeta}^\star_\text{fus}&=-\Big(\sum_{i=1}^N \sum_{j=1}^N  \vect{\mathcal{J}}_r^{-\top}(\vect{\zeta}_j)\vect{G}_{ij}^\top\vect{\mathcal{J}}_r^{-1}(\vect{\zeta}_i)\\\nonumber&~~~~~~~+  \vect{\mathcal{J}}_r^{-\top}(\vect{\zeta}_i)\vect{G}_{ij}\vect{\mathcal{J}}_r^{-1}(\vect{\zeta}_j)\Big)^{-1}\\&\sum_{i=1}^N \sum_{j=1}^N\vect{\mathcal{J}}_r^{-\top}(\vect{\zeta}_j)\vect{G}_{ij}^\top\vect{\zeta}_i+\vect{\mathcal{J}}_r^{-\top}(\vect{\zeta}_i)\vect{G}_{ij}\vect{\zeta}_j,\\\nonumber
\vect{P}_{\vect{\zeta}\vect{\zeta}}^\star&= \Big(\sum_{i=1}^N \sum_{j=1}^N  \vect{\mathcal{J}}_r^{-\top}(\vect{\zeta}_j)\vect{G}_{ij}^\top\vect{\mathcal{J}}_r^{-1}(\vect{\zeta}_i)\\&~~~~~+  \vect{\mathcal{J}}_r^{-\top}(\vect{\zeta}_i)\vect{G}_{ij}\vect{\mathcal{J}}_r^{-1}(\vect{\zeta}_j)\Big)^{-1},  
\end{align}
 where $\vect{\zeta}_i=\left[\log(\Bvect{\mathcal{X}}_i^{-1}\Bvect{\mathcal{X}}_s)\right]_\vee$ and
\begin{align}\nonumber
    \vect{G}\!=\!\left[\begin{smallmatrix}
        \vect{P}_{11}&\cdots&\vect{P}_{1n}\\\vdots&\ddots&\vdots\\\vect{P}_{1n}^\top&\cdots&\vect{P}_{nn}\end{smallmatrix}\right]^{-1}=\left[\begin{smallmatrix}
        \vect{G}_{11}&\cdots&\vect{G}_{1N}\\
        \vdots&\ddots&\vdots\\\vect{G}_{N1}&\cdots&\vect{G}_{NN}
    \end{smallmatrix}\right]\in\mathbb{R}^{\mathfrak{m}n\times \mathfrak{m}n}.
\end{align}


\section{Dual-View Target Object Localization}\label{sec-loc}
Here,  we first develop the core of our estimation framework which is a Lie group-based adaptive EKF inspired by the original EKF developed in~\cite{bourmaud2013discrete}. We then formulate the tracking problem in terms of process and measurement models consistent with the structure of the proposed filter. Finally, we present the decentralized estimation framework for dual-view pose and velocity estimation of the target object.

\subsection{Adaptive EKF on Matrix Lie Groups}\label{sec: ekf}
We define a stochastic member on an arbitrary matrix Lie group ${\mathcal{G_X}}$ as $\vect{\mathcal{X}}=\Bvect{\mathcal{X}}\exp([\vect{\zeta}]_\wedge)$  
where   $\Bvect{\mathcal{X}}$ denotes the deterministic part and $\vect{\zeta}\sim\mathcal{N}(\vect{0}_{\mathfrak{m}\times 1}, \vect{P})$ is a zero-mean Gaussian vector with the covariance $\vect{P}\in\mathbb{R}^{\mathfrak{m}\times \mathfrak{m}}$. Let a discrete-time system on ${\mathcal{G_X}}$  with noisy measurements from a sensor be:
\begin{align}\label{eq: process model}
 & \vect{\mathcal{X}}{\scriptstyle (k+1)}=\!\vect{\mathcal{X}}{\scriptstyle (k)}\exp(\left[\vect{f}(\vect{\mathcal{X}}{\scriptstyle (k)}, \vect{u}{\scriptstyle (k)})\!+\!\vect{w}{\scriptstyle (k)}\right]_\wedge)  ,\\\label{eq: process model2}
  &\vect{\mathcal{Z}}{\scriptstyle (k)}\!=\!\vect{h}(\vect{\mathcal{X}}{\scriptstyle (k)})\exp(\left[{\vect{m}{\scriptstyle (k)}}\right]_{\wedge}), 
\end{align}
where the measurement signal $\vect{\mathcal{Z}}{\scriptstyle (k)}\in{\mathcal{G_Z}}$ is on a $q$-dimensional matrix Lie group according to the model $\vect{h}: {\mathcal{G_X}}\rightarrow{\mathcal{G_Z}}$ and  $\vect{u}{\scriptstyle (k)}\in\mathbb{R}^r$ is the control input to the process whose model is described by the function
$\vect{f}: {\mathcal{G_X}}\times\mathbb{R}^r\rightarrow \mathbb{R}^\mathfrak{m}$. The covariance of the normal white noise sequences $\vect{w}{\scriptstyle (k)}\sim\mathcal{N}(\vect{0}_{\mathfrak{m}\times 1},\vect{Q}{\scriptstyle (k)}) $ and $\vect{m}{\scriptstyle (k)}\sim\mathcal{N}(\vect{0}_{q\times 1},\vect{R}{\scriptstyle (k)})$ are $\vect{Q}{\scriptstyle (k)}\in \mathbb{R}^{\mathfrak{m}\times \mathfrak{m}}$ and $\vect{R}{\scriptstyle (k)}\in\mathbb{R}^{q\times q}$, respectively. We assume that the measurement and process noise signals are independent, i.e., $\mathbb{E}[\vect{m}{\scriptstyle (k)}\vect{w}^\top{\scriptstyle (l)}]=\vect{0}_{q\times \mathfrak{m}}$.

Given the mean and covariance at the time ${\scriptstyle k-1}$, represented as $\Bvect{\mathcal{X}}{\scriptstyle (k-1|k-1)}$ and $\vect{P}{\scriptstyle (k-1|k-1)}$, respectively, the prediction step of the EKF calculates the prior estimates as~\cite{bourmaud2013discrete}:
\begin{align}\label{eq: mean and cov in propagation}
 \Bvect{\mathcal{X}}{\scriptstyle (k|k-1)}&=\Bvect{\mathcal{X}}{\scriptstyle (k-1|k-1)}\exp(\left[\Bvect{f}{\scriptstyle (k-1)}\right]_\wedge),\\\label{eq: mean and cov in propagation 2}\nonumber 
 \vect{P}{\scriptstyle (k|k-1)}&=\vect{\mathcal{F}}{\scriptstyle (k-1)}\vect{P}{\scriptstyle (k-1|k-1)}\vect{\mathcal{F}}^\top{\scriptstyle (k-1)}\\&+\vect{\mathcal{J}}_r(\Bvect{f}{\scriptstyle (k-1)})\vect{Q}{(\scriptstyle k-1)}\vect{\mathcal{J}}_r^\top(\Bvect{f}{\scriptstyle (k-1)}),
\end{align}
where $\Bvect{f}({\scriptstyle k-1})=\vect{f}(\Bvect{\mathcal{X}}({\scriptstyle k-1|k-1}), \vect{u}({\scriptstyle k-1}))$ and
\begin{align}\label{eq: matrix F}
 & \vect{\mathcal{F}}{\scriptstyle (k)}=\vect{Ad}_{\exp(\left[-\Bvect{f}{\scriptstyle (k-1)}\right]_\wedge)}+ \vect{\mathcal{J}}_r\big(\Bvect{f}{\scriptstyle (k-1)}\big)\vect{\mathcal{D}}{\scriptstyle (k-1)},\\\label{eq: matrix D}
&\vect{\mathcal{D}}{\scriptstyle (k-1)}\!=\!\frac{\partial}{\partial \vect{\zeta}}\vect{f}(\Bvect{\mathcal{X}}{\scriptstyle (k-1|k-1)}\exp([\vect{\zeta}]_\wedge),\vect{u}{\scriptstyle (k-1)})\Big|_{\vect{\zeta}=\vect{0}}. 
 \end{align}
At time step ${\scriptstyle k}$, the update step is executed based on 
 \begin{align}
    \Bvect{\mathcal{X}}{\scriptstyle (k|k)}&=\Bvect{\mathcal{X}}{\scriptstyle (k|k-1)}\exp\big(\big[\vect{K}{\scriptstyle (k)}\vect{\nu}{\scriptstyle (k)}\big]_\wedge\big),\\\nonumber
    \vect{P}{\scriptstyle (k|k)}&=\vect{\mathcal{J}}_r\big(\vect{K}{\scriptstyle (k)}\vect{\nu}{\scriptstyle (k)}\big)\big(\vect{I}_{\mathfrak{m}}-\vect{K}{\scriptstyle (k)}\vect{H}{\scriptstyle (k)}\big)\vect{P}{\scriptstyle (k|k-1)}\\&~~~\vect{\mathcal{J}}_r^\top\big(\vect{K}{\scriptstyle (k)}\vect{\nu}{\scriptstyle (k)}\big),
\end{align}
where the innovation $\vect{\nu}{\scriptstyle (k)}$, the EKF gain, and matrix $\vect{\mathcal{H}}{\scriptstyle (k)}$ are
\begin{align}
    \vect{\nu}{\scriptstyle (k)}&=\big[\log\big(\vect{h}^{-1}(\Bvect{\mathcal{X}}{\scriptstyle (k|k-1)})\vect{\mathcal{Z}}{\scriptstyle (k)}\big)\big]_\vee,\\\label{eq: kalman gain}
    \vect{K}{\scriptstyle (k)}&\!=\!\vect{P}{\scriptstyle (k|k-1)}\vect{\mathcal{H}}^\top\!{\scriptstyle (k)}\big(\vect{\mathcal{H}}{\scriptstyle (k)}\vect{P}{\scriptstyle (k|k-1)}\vect{\mathcal{H}}^\top\!{\scriptstyle (k)}\!+\!\vect{R}{\scriptstyle (k)}\big)^{-1}\!,\!\\\label{eq: matrix H}
    \vect{\mathcal{H}}{\scriptstyle (k)}&\!=\!\frac{\partial}{\partial\vect{\zeta}}\!\big[\log\big(\scriptstyle{\vect{h}^{-1}(\Bvect{\mathcal{X}}(k|k-1))\vect{h}(\Bvect{\mathcal{X}}(k|k-1)\exp([\vect{\zeta}]_\wedge))\big)\big]_\vee\!\Big|_{\vect{\zeta}=\vect{0}}}.
\end{align}

Achieving satisfactory performance with the EKF and the subsequent fusion process requires careful tuning of the process and measurement noise covariance matrices, $\vect{Q}{\scriptstyle (k)}$ and $\vect{R}{\scriptstyle (k)}$~\cite{mehra1970adaptive}. We propose the adaptive tuning of $\vect{Q}{\scriptstyle (k)}$ and $\vect{R}{\scriptstyle (k)}$ at each time step based on innovation-based and exponentially weighted moving average techniques. To this end, inspired by~\cite{mohamed1999adaptive} we compute the following  residual vector at each time step:
\begin{align}
  \vect{\mu}{\scriptstyle (k)}&=\big[\log\big(\vect{h}^{-1}(\Bvect{\mathcal{X}}{\scriptstyle (k|k)})\vect{\mathcal{Z}}{\scriptstyle (k)}\big)\big]_\vee.
\end{align}
The process and measurement noise covariances are then recursively estimated using exponentially weighted moving average structure as follows:
\begin{align*}
  \vect{Q}{\scriptstyle (k)}&\!=\!f_{\vect{Q}}\vect{Q}{\scriptstyle (k-1)}+(1\!-\!f_{\vect{Q}})(\vect{K}{\scriptstyle (k)}\vect{\nu}{\scriptstyle (k)}\vect{\nu}^\top{\scriptstyle (k)}\vect{K}^\top{\scriptstyle (k)}), \\
 \vect{R}{\scriptstyle (k)}&\!=\!f_{\vect{R}}\vect{R}{\scriptstyle (k-1)}\!+\!(\!1\!-\!f_{\vect{R}})(\vect{\mu}{\scriptstyle (k)}\!\vect{\mu}^\top\!{\scriptstyle (k)}\!+\!\vect{\mathcal{H}}{\scriptstyle (k)}\!\vect{P}{\scriptstyle (k|k-1)}\vect{\mathcal{H}}^\top\!{\scriptstyle (k)}),  
\end{align*}
where hyperparameters $0 \leq f_{\vect{Q}}\leq 1$ and $0 \leq f_{\vect{R}}\leq 1$ are forgetting factors. When $f_{\vect{Q}}=f_{\vect{R}}=1$ the adaptive EKF reduces to the original EKF with constant noise covariances.
 
\color{black}

\subsection{Uncertain Dynamics of the Target Object}
As shown in Fig.\ref{fig::general_manipulator}, we assume the grasping frame $\{G\}$ is known and fixed on a target object that moves with constant velocity with respect to the base frame $\{B\}$ of the manipulator. Let $\vect{\mathcal{T}}^{BG}(t)=\left[\begin{smallmatrix}
           \vect{\mathcal{R}}^{BG}{\scriptstyle (t)}&\vect{p}_G^B{\scriptstyle (t)}\\\vect{0}_{1\times3}&1
   \end{smallmatrix}\right]\in\mathbb{SE}(3)$ represent the transformation between the grasping coordinate frame attached on the target object and the spatial base link of the manipulator. Here, $\vect{\mathcal{R}}^{BG}{\scriptstyle (t)}\in\mathbb{SO}(3)$ is the rotation matrix of frame $\{G\}$ relative to frame $\{B\}$, while  $\vect{p}_G^B{\scriptstyle (t)}\in\mathbb{R}^3$ denotes the  position of point $G$ relative to frame $\{B\}$ and expressed in frame $\{B\}$. The noise-corrupted constant velocity motion of the target object is defined by:
\begin{align}
 \dot{\vect{\mathcal{T}}}^{BG}{\scriptstyle (t)}&={\vect{\mathcal{T}}}^{BG}{\scriptstyle (t)}[\vect{\Omega}{\scriptstyle (t)}]_\wedge,\\   \dot{\vect{\Omega}}{\scriptstyle (t)}&=\vect{n}{\scriptstyle (t)},
\end{align}
where $\vect{\Omega}{\scriptstyle (t)}=\left[\begin{smallmatrix}
    \vect{\omega}^{BG}{\scriptstyle (t)}\\\vect{v}_G^B{\scriptstyle (t)}
\end{smallmatrix}\right]\in\mathbb{R}^6$, with $\vect{\omega}^{BG}{\scriptstyle (t)}\in\mathbb{R}^3$ being the angular velocity of frame $\{G\}$ relative to frame $\{B\}$ and $\vect{v}_G^B{\scriptstyle (t)}\in\mathbb{R}^3$ representing the velocity of point $G$ relative to frame $\{B\}$ and expressed in frame $\{G\}$. Moreover, $\vect{n}{\scriptstyle (t)}=\left[\begin{smallmatrix}
\vect{n}_\omega{\scriptstyle (t)}\\\vect{n}_v{\scriptstyle (t)}\end{smallmatrix}\right]\in\mathbb{R}^6$ is a zero-mean Gaussian noise with covariance $\vect{Q}_n{\scriptstyle (t)}\in\mathbb{R}^{6\times 6}$ where $\vect{n}_\omega{\scriptstyle (t)}\sim\mathcal{N}(\vect{0}_{3\times1},\vect{Q}_{n_w}{\scriptstyle (t)})$ and $\vect{n}_v{\scriptstyle (t)}\sim\mathcal{N}(\vect{0}_{3\times1},\vect{Q}_{n_v}{\scriptstyle (t)})$ respectively denote the rotational and translational noise vectors. In order to simplify the analysis and provide a more elegant mathematical formulation, we unify these equations into a single by introducing an augmented matrix Lie group $\mathcal{G}_{\mathcal{X}}\coloneq\mathbb{SE}(3)\times\mathbb{R}^3\times\mathbb{R}^3$ whose members $\vect{\mathcal{X}}{\scriptstyle (t)}\in\mathbb{R}^{9\times9}$ are 
\begin{align}\nonumber
\vect{\mathcal{X}}{\scriptstyle (t)}\!=\!\begin{bmatrix}
   \begin{bmatrix}
       \vect{\mathcal{R}}^{BG}{\scriptstyle (t)}&\vect{p}_G^B{\scriptstyle (t)}\\\vect{0}_{1\times3}&1
   \end{bmatrix}&\vect{0}_{4\times5}\\\vect{0}_{5\times4}&\begin{bmatrix} \vect{I}_{3}&\vect{\omega}^{BG}{\scriptstyle (t)}&\vect{v}_G^B{\scriptstyle (t)}\\\vect{0}_{1\times3}&1&0\\\vect{0}_{1\times3}&0&1\end{bmatrix}
\end{bmatrix}.
\end{align}
The augmented system is then defined as:
\begin{align}
 \dot{\vect{\mathcal{X}}}{\scriptstyle (t)}= \vect{\mathcal{X}}{\scriptstyle (t)} \left[\vect{f}(\vect{\mathcal{X}}{\scriptstyle (t)})\!+\!\vect{w}{\scriptstyle (t)}\right]_\wedge.
\end{align}

The matrix $\big[\vect{f}\big(\vect{\mathcal{X}{\scriptstyle (t)}}\big)+\vect{w}{\scriptstyle (t)}\big]_\wedge=\vect{\mathcal{X}}^{-1}{\scriptstyle (t)}\dot{\vect{\mathcal{X}}}{\scriptstyle (t)}\in\mathfrak{g}_{\mathcal{X}}$, where $\mathfrak{g}_{\mathcal{X}}$ denotes the Lie algebra of $\mathcal{G_X}$ and its members have the following form:
\begin{align}\nonumber
\begin{bmatrix}
   \begin{bmatrix}
       \left[\vect{\omega}^{BG}{\scriptstyle (t)}\right]_\times&\vect{v}_G^B{\scriptstyle (t)}\\\vect{0}_{1\times3}&0
   \end{bmatrix}&\vect{0}_{4\times5}\\\vect{0}_{5\times4}&\begin{bmatrix} \vect{0}_{3}&\vect{n}_{\omega}{\scriptstyle (t)}&\vect{n}_v{\scriptstyle (t)}\\\vect{0}_{1\times3}&0&0\\\vect{0}_{1\times3}&0&0\end{bmatrix}
\end{bmatrix}\in\mathbb{R}^{9\times 9}.
\end{align}
Therefore, the vector $\vect{f}\big(\vect{\mathcal{X}}{\scriptstyle (t)}\big)$ and $\vect{w}{\scriptstyle (t)}$ are calculated as
\begin{align}
 \vect{f}\big(\vect{\mathcal{X}}{\scriptstyle (t)}\big)\!=\!\begin{bmatrix}
     \vect{\omega}^{BG}{\scriptstyle (t)}\\\vect{v}_G^B{\scriptstyle (t)}\\
     \vect{0}_{1\times3}\\
     \vect{0}_{1\times3}\\
     \vect{0}_{1\times3}
 \end{bmatrix},~\vect{w}{\scriptstyle (t)}\!=\!\begin{bmatrix}
     \vect{0}_{1\times3}\\
     \vect{0}_{1\times3}\\
     \vect{0}_{1\times3}\\
     \vect{n}_{\omega}{\scriptstyle (t)}\\
     \vect{n}_v{\scriptstyle (t)}
 \end{bmatrix}\sim\mathcal{N}(\vect{0}_{15\times 1},\vect{Q}{\scriptstyle (t)}).  
\end{align}
Here, the process noise covariance matrix is $\vect{Q}{\scriptstyle (t)}=\operatorname{diag}([\vect{0}_{3},\vect{0}_{3},\vect{0}_{3},\vect{Q}_n{\scriptstyle (t)}])\in\mathbb{R}^{15\times15}$.
The Euler method is then employed to perform discretization
of this continuous-time stochastic system:
\begin{align}
\vect{\mathcal{X}}\!{\scriptstyle (k+1)}\!=\!\vect{\mathcal{X}}\!{\scriptstyle (k)}\exp\big(\!\Delta t [\vect{f}({ \vect{\mathcal{X}}\!\scriptstyle(k)}]_\wedge\!+\!\sqrt{\Delta t}[\vect{w}{\scriptstyle (k)}]_\wedge\big),
\end{align}  
where $\Delta t=t_{k+1}-t_k$ represents the discretization interval, and $\vect{w}{\scriptstyle (k)}$  represents the acceleration noise that perturbs the constant-velocity motion. This discretized system is used during the prediction step of the EKF. Moreover, 
With this process  model the matrix $\vect{\mathcal{D}}$ is calculated as:
\begin{align*}
\vect{\mathcal{D}}{\scriptstyle (k)}=\begin{bmatrix}
    \vect{0}_{6\times 9}&\vect{I}_{6}\\\vect{0}_{9}&\vect{0}_{9\times 6}
\end{bmatrix}.
\end{align*}

\subsection{Camera Pose Measurements}
\begin{figure}[htbp]
\centerline{\includegraphics[width=0.45\textwidth]{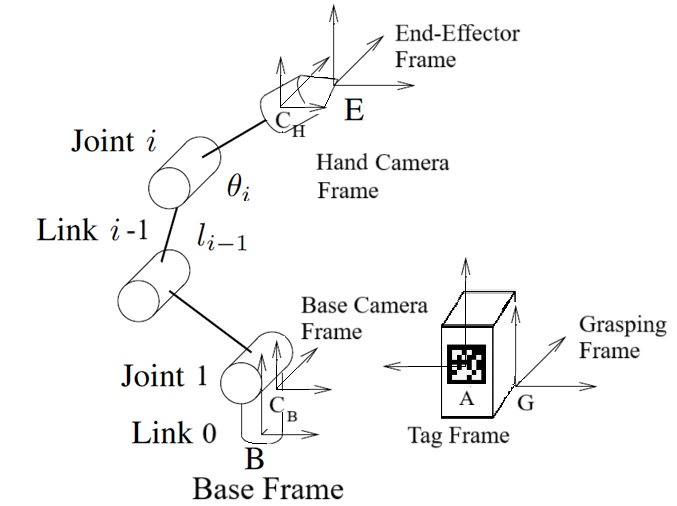}}
\caption{Coordinate frames of the  manipulator, cameras, and the target objects.}
\label{fig::general_manipulator}
\end{figure}

As illustrated in Fig.~\ref{fig::general_manipulator} there are two cameras installed on the robot, one is eye-in-hand (hand camera) with coordinate frame $\{C_H\}$ and the other one is eye-to-hand (base camera) installed on the base with coordinate frame $\{C_B\}$. We can attach one tag to be seen by both cameras, or one tag for each camera which has more flexibility. Let $\{A_B\}$ be the coordinate frame of the tag seen by the base camera and $\{A_H\}$ be the coordinate frame of the tag seen by the hand camera. Then pseudo-poses captured by each camera can be expressed as:
\begin{align}\label{eq::hand_meas}
  \vect{\mathcal{Z}}_{\text{hand}}\coloneq\vect{\mathcal{T}}^{BG}_{\text{hand}}&=\vect{\mathcal{T}}^{BE}\vect{\mathcal{T}}^{EC_H}\vect{\mathcal{T}}^{C_HA_H}\vect{\mathcal{T}}^{A_HG},\\\label{eq::base_meas}
  \vect{\mathcal{Z}}_{\text{base}}\coloneq\vect{\mathcal{T}}^{BG}_{\text{base}}&=\vect{\mathcal{T}}^{BC_B}\vect{\mathcal{T}}^{C_BA_B}\vect{\mathcal{T}}^{A_BG}.
\end{align}
Note that the transformations $\vect{\mathcal{T}}^{A_HG}$, $\vect{\mathcal{T}}^{A_BG}$, $\vect{\mathcal{T}}^{EC_H}$, $\vect{\mathcal{T}}^{BC_B}$ are known and constant, while the transformations $\vect{\mathcal{T}}^{C_HA_H}$ and $\vect{\mathcal{T}}^{C_BA_B}$ are provided by the tag detection package using, respectively, the hand and base cameras. Moreover, the transformation $\vect{\mathcal{T}}^{BE}$ comes from the forward kinematics of the manipulator. The forward kinematics of a manipulator is a mapping from the joint space, represented by all possible joint variables $\vect{\theta}=[\theta_1,\cdots,\theta_n]^\top$, to the special Euclidean group $\mathbb{SE}(3)$, which determines the configuration of the end-effector frame $\{E\}$ relative to the base frame $\{B\}$ being $\vect{\mathcal{T}}^{BE} $, as illustrated in Fig.~\ref{fig::general_manipulator}.

Using the product of exponentials formula, the manipulator forward kinematics is formulated as follows~\cite{murray2017mathematical}:
\begin{align}
\vect{\mathcal{T}}^{BE}(\vect{\theta})=\exp(\left[\vect{\zeta}_1\right]_\wedge\theta_1)\cdots\exp(\left[\vect{\zeta}_n\right]_\wedge\theta_n)\vect{\mathcal{T}}^{BE}(\vect{0}).
\end{align}
The member $\vect{\mathcal{T}}^{BE}(\vect{0})\in \mathbb{SE}(3)$ is the reference configuration of the manipulator when all joint variables are zero, and for the $i^{th}$ revolute joint  $\vect{\zeta}_i=\begin{bmatrix}
     \vect{\omega}_i^\top&(-\vect{\omega}_i\times q_i)^\top
 \end{bmatrix}^\top$, where $\vect{\omega}_i\in\mathbb{R}^3$ is a unit vector in the direction of the joint axis and $q_i\in\mathbb{R}^3$ is an arbitrary point on the axis~\cite{murray2017mathematical}.

The constructed pseudo-poses in \eqref{eq::hand_meas}-\eqref{eq::base_meas} are noise corrupted and we assume the noise follows a zero-mean Gaussian process. Therefore, we can consider the following measurement model for both cameras in the update step of the EKF:
\begin{align}
  \vect{\mathcal{Z}}_i{\scriptstyle (k)}=\vect{\mathcal{T}}^{BG}{\scriptstyle (k)}\exp(\left[{\vect{m}}_i{\scriptstyle (k)}\right]_{\wedge}), i\in\{\text{base},\text{hand}\},
\end{align}
where $\vect{m}_i{\scriptstyle (k)}\sim\mathcal{N}(\vect{0}_{6\times 1},\vect{R}_i{\scriptstyle (k)})$ with $\vect{R}_i{\scriptstyle (k)}\in\mathbb{R}^{6\times6}$. With this measurement models the matrix  $\vect{\mathcal{H}}$ is calculated as:
\begin{align*}
\vect{\mathcal{H}}=\begin{bmatrix} \vect{I}_{6}&\vect{0}_{6\times 9}\end{bmatrix}.
\end{align*}

\subsection{ Pose and velocity Estimation and Fusion Framework}


The flowchart illustrating the estimation and fusion framework is shown in Fig.~\ref{fig: estimation flowchart}. The robot runs two independent filters in a decentralized manner. Both filters share an identical prediction step based on the constant motion dynamics and the state is propagated on the manifold $\mathbb{SE}(3) \times \mathbb{R}^3 \times \mathbb{R}^3$. Each filter performs a measurement update on the manifold $\mathbb{SE}(3)$ based on the availability of pseudo-pose measurements from its corresponding sensor. The hand camera update can only be executed when both the forward kinematics, $\vect{\mathcal{T}}^{BE}{\scriptstyle (k)}$, and the camera measurements, $\vect{\mathcal{T}}^{C_HA_H}{\scriptstyle (k)}$, are available at time step~$k$ while the base camera update only depends on  the availability of $\vect{\mathcal{T}}^{C_BA_B}{\scriptstyle (k)}$. If this information is not available, the posterior state is set equal to the prior, and the system proceeds to the next time step using the process model dynamics. The estimated states, denoted by $\Bvect{\mathcal{X}}_h{\scriptstyle (k|k)}$ and $\Bvect{\mathcal{X}}_b{\scriptstyle (k|k)}$, are subsequently fused in a  fusion module using a correlation-aware fusion strategy developed in~\cite{zarei2024consistent}. For this EKF-based estimation framework, the cross-covariance matrix between correlated estimates is calculated from the following recursion:  
\begin{align*}
&\nonumber\vect{P}_{bh}{\scriptstyle (k|k)}\!=\!\\&
\big(\vect{I}_{15}\!-\!\vect{K}_b{\scriptstyle (k)}\vect{\mathcal{H}}_b{\scriptstyle (k)}\big)\Big(\vect{\mathcal{F}}_b{\scriptstyle(k-1)}\vect{P}_{bh}{\scriptstyle( k-1|k-1)}\vect{\mathcal{F}}_h^\top{\scriptstyle( k-1)}\\&+\vect{\mathcal{J}}_r(\Bvect{f}_b{\scriptstyle( k-1)})\vect{Q}{\scriptstyle( k-1)}\vect{\mathcal{J}}_r^\top(\Bvect{f}_h{\scriptstyle( k-1)})\!\Big)\big(\vect{I}_{15}\!-\!\vect{K}_h{\scriptstyle (k)}\vect{\mathcal{H}}_h{\scriptstyle (k)}\big)\!^\top.
\end{align*}

The initial condition of this recursion is $\vect{P}_{bh}{\scriptstyle (0|0)}=\vect{0}_{15}$.




\begin{figure}[h!]
\centering
\scalebox{0.6}{
\begin{tikzpicture}[node distance=2.5cm, 
    process/.style={rectangle, draw, minimum width=2.5cm, minimum height=1cm, text centered},
    decision/.style={diamond, draw, minimum width=2.5cm, minimum height=1cm, text centered, aspect=2},
    arrow/.style={->, thick}]

\node (initial) [process] {\parbox{3cm}{\centering Initial Conditions\\$\Bvect{\mathcal{X}}_{b,h}{\scriptstyle(0|0)}, \vect{P}_{b,h}{\scriptstyle(0|0)}$}};
\node (prediction) [process, below=2cm of initial] {\parbox{3cm}{\centering EKFs Predictions}};

\node (hand_decide) [decision, below left=1.5cm and 2cm of prediction] {\parbox{2cm}{\centering $\vect{\mathcal{Z}}_{\text{hand}}{\scriptstyle(k)}$ \\ Available?}};
\node (base_decide) [decision, below right=1.5cm and 2cm of prediction] {\parbox{2cm}{\centering $\vect{\mathcal{Z}}_{\text{base}}{\scriptstyle(k)}$ \\ Available?}};

\node (hand_update) [process, below=2cm of hand_decide] {\parbox{3cm}{\centering Hand EKF Update\\
$\Bvect{\mathcal{X}}_h{\scriptstyle(k|k)}, \vect{P}_h{\scriptstyle(k|k)}$\\
$\vect{Q}_h{\scriptstyle(k)}, \vect{R}_h{\scriptstyle(k)}$}};

\node (base_update) [process, below=2cm of base_decide] {\parbox{3cm}{\centering Base EKF Update\\
$\Bvect{\mathcal{X}}_b{\scriptstyle(k|k)}, \vect{P}_b{\scriptstyle(k|k)}$\\
$\vect{Q}_b{\scriptstyle(k)}, \vect{R}_b{\scriptstyle(k)}$}};

\node (fusion) [process, below=3.5cm of prediction] {\parbox{3cm}{\centering Fusion\\
$\Bvect{\mathcal{X}}_\text{fuse}{\scriptstyle(k|k)}, \vect{P}_\text{fuse}{\scriptstyle(k|k)}$}};

\node (noupdate_base) [process, right=4cm of base_decide] {\parbox{3cm}{\centering No Update\\
$\Bvect{\mathcal{X}}_b{\scriptstyle(k|k)}=\Bvect{\mathcal{X}}_b{\scriptstyle(k|k-1)}$\\
$\vect{P}_b{\scriptstyle(k|k)}=\vect{P}_b{\scriptstyle(k|k-1)}$}};

\node (noupdate_hand) [process, left=4cm of hand_decide] {\parbox{3cm}{\centering No Update\\
$\Bvect{\mathcal{X}}_h{\scriptstyle(k|k)}=\Bvect{\mathcal{X}}_h{\scriptstyle(k|k-1)}$\\
$\vect{P}_h{\scriptstyle(k|k)}=\vect{P}_h{\scriptstyle(k|k-1)}$}};

\draw [arrow] (initial) -- (prediction);
\draw [arrow] (prediction.south) -- ++(-2,-0.5) -- (hand_decide);
\draw [arrow] (prediction.south) -- ++(2,-0.5) -- (base_decide);

\draw [arrow] (hand_decide.south) -- (hand_update);
\draw [arrow] (base_decide.south) -- (base_update);

\draw [arrow] (hand_update.south) -- ++(0,-1) -- (fusion);
\draw [arrow] (base_update.south) -- ++(0,-1) -- (fusion);

\draw [arrow] (hand_decide.west) -- ++(-2,0) -- (noupdate_hand.east);
\draw [arrow] (base_decide.east) -- ++(2,0) -- (noupdate_base.west);

\draw [arrow] (noupdate_hand.north) -- ++(0,1) -- ++(1.5,0) -- (prediction.west);
\draw [arrow] (noupdate_base.north) -- ++(0,1) -- ++(-1.5,0) -- (prediction.east);

\draw [arrow] (fusion.south) -- ++(0,-1); 

\end{tikzpicture}
}
\caption{Estimation and fusion framework of the manipulator.}
\label{fig:estimation_flowchart}
\end{figure}
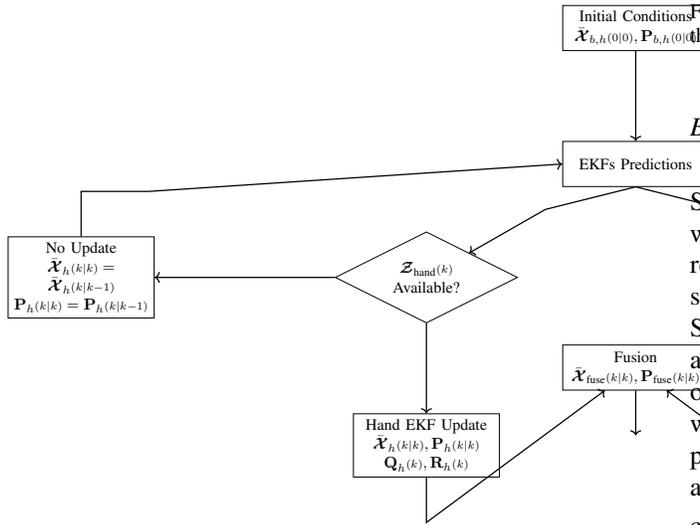

\section{Experiments and Results}\label{sec-exp}

\subsection{Setup}
\begin{figure}[!h]
  \unitlength=0.5in
   \centering 
    \subfloat[]{
   \includegraphics[width=1.5in]{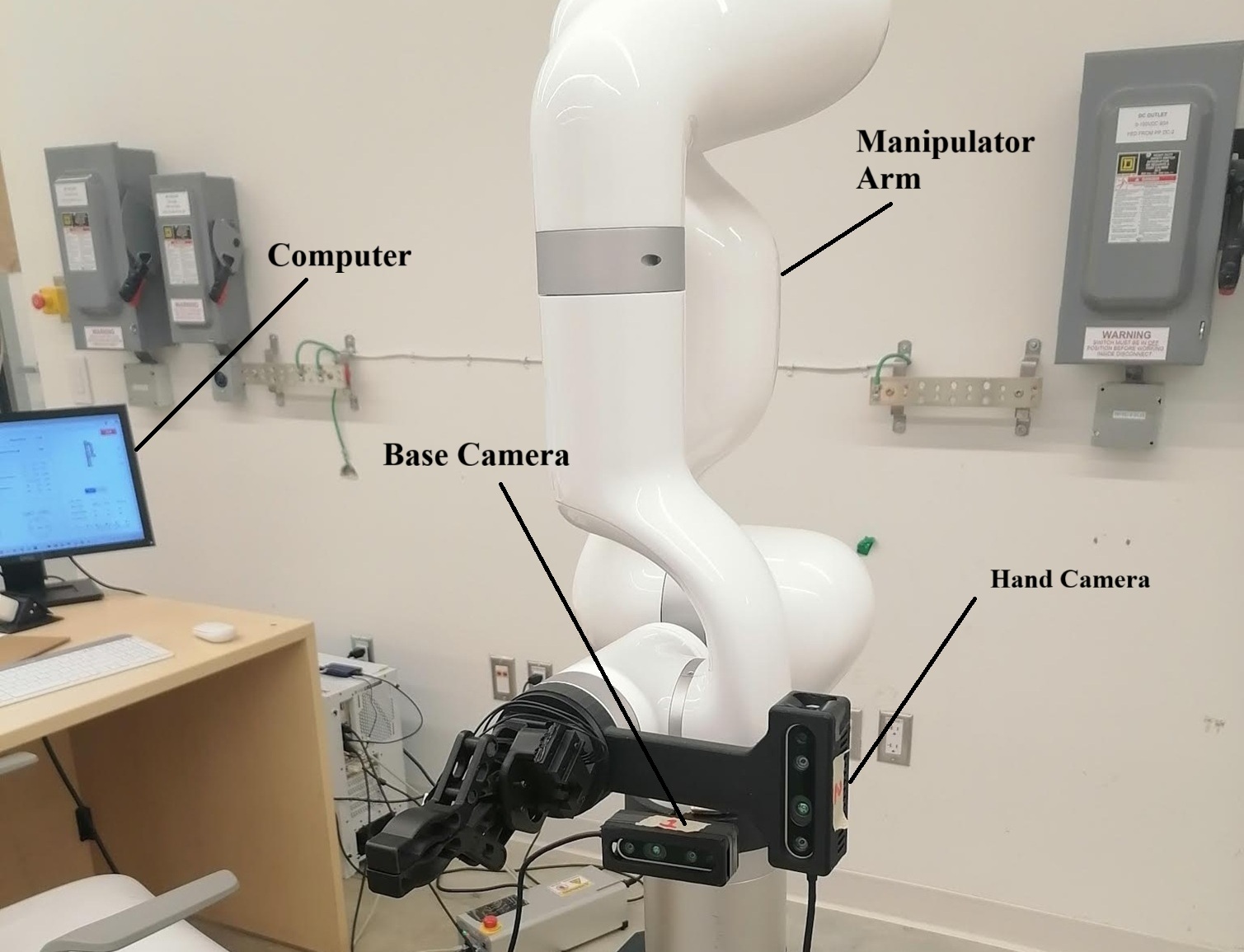}
  }~
      \subfloat[]{ \includegraphics[width=1.5in]{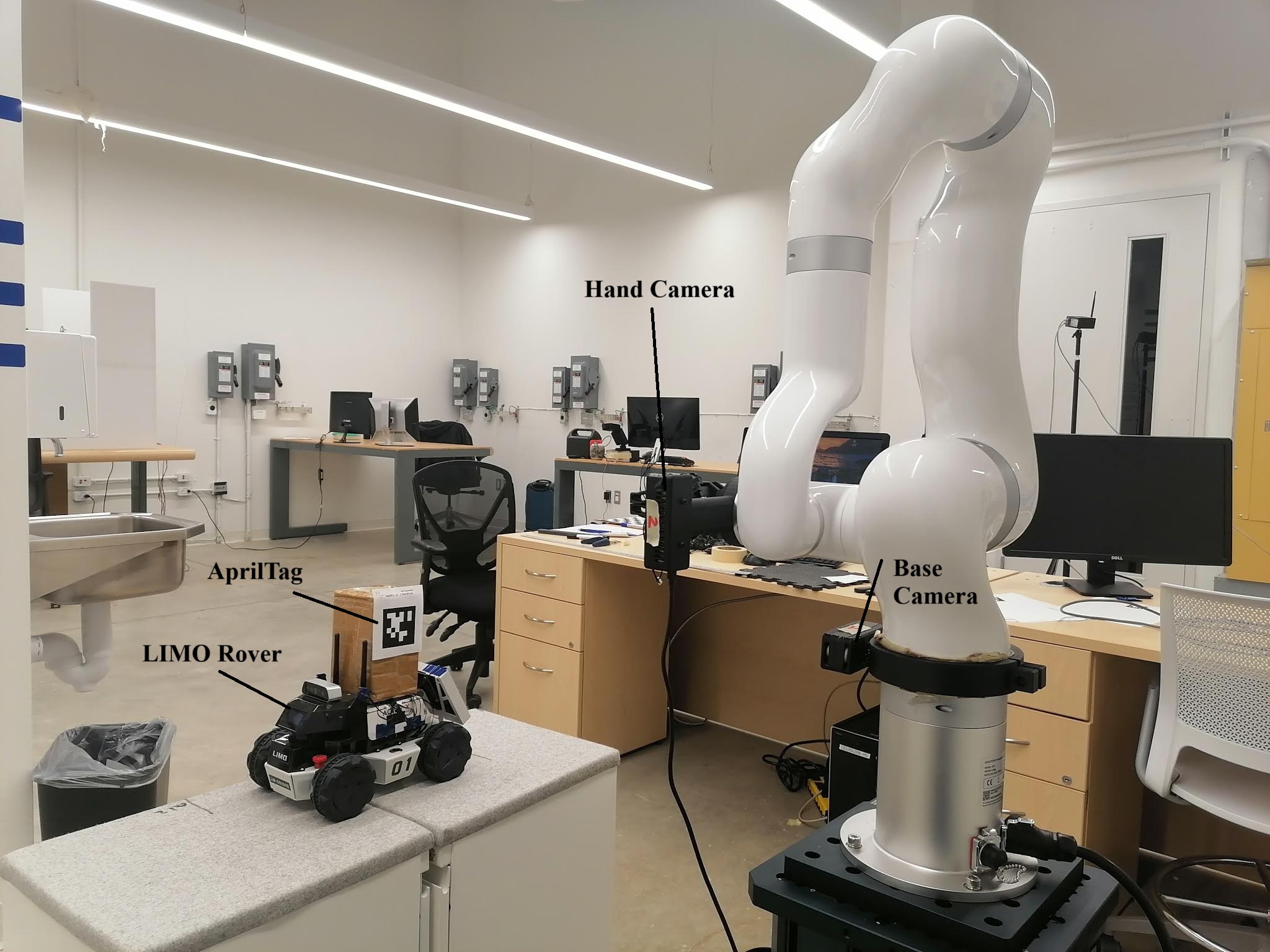}
       
   }  
  \\
\subfloat[]{ \includegraphics[width=1.5in]{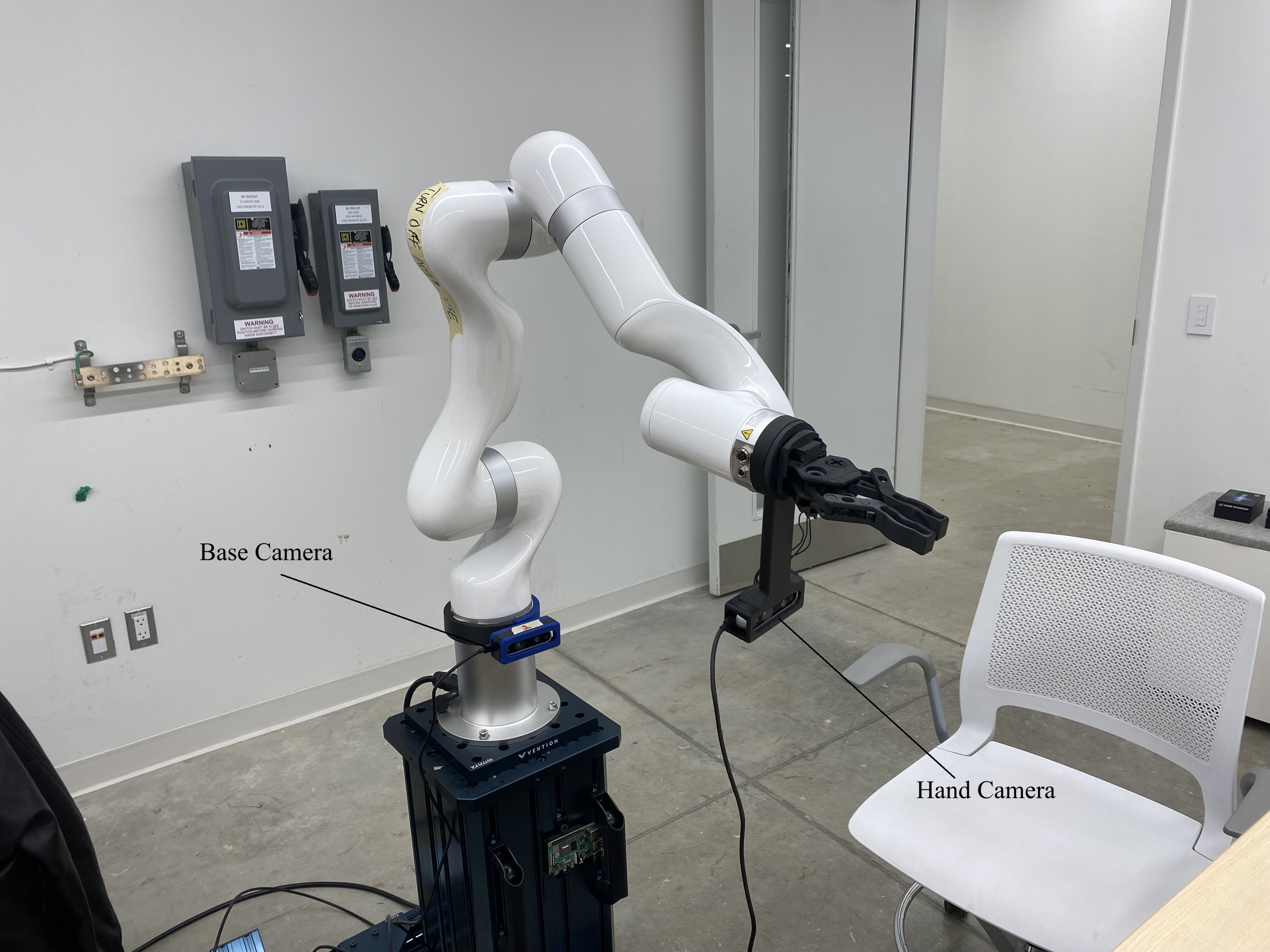}
       
   }~\subfloat[]{ \includegraphics[width=1.5in]{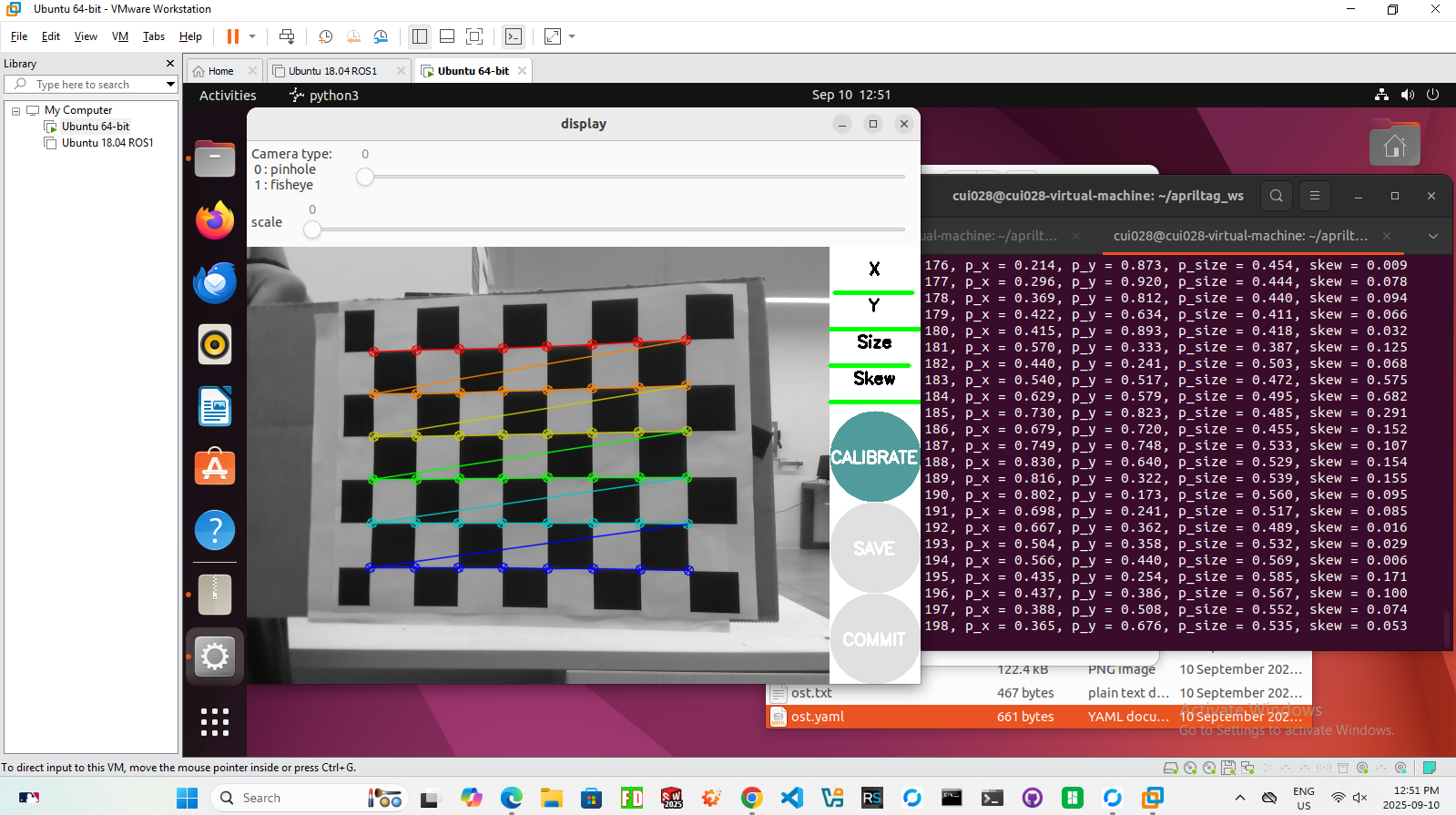}
       
   }
  \\
   \caption{The setup: (a) front view of the cameras, (b) the rover carrying the object, (c) the arm following the target,  and (d) camera calibration. }\label{fig: expriment setup}
\end{figure}

In the experimental study, we set up a controlled and representative testing environment consisting of a LIMO rover carrying a cubic target object marked with an AprilTag, a UFactory xArm 6 robotic manipulator ($850mm$ reach, 6 DOF, $\pm0.1mm$ repeatability), two Intel RealSense D435 depth cameras~\cite{realSenseCam2025}, and a computer (See Fig.\ref{fig: expriment setup}).  Camera calibration was performed using the \emph{camera-calibration} ROS package~\cite{CamCalib2025} and involved capturing 147 images of an $8\times6$
checkerboard with a square size of $3cm$. To establish the localization ground truth, we used the odometry data published by the LIMO rover~\cite{limo2025}, which is sufficiently accurate for  constant velocity motion. Using the \emph{apriltag-ros} 
package~\cite{apriltag2025}, the algorithm computes the tag's 3-D pose relative to  the base and hand camera frames. In our experiments, we used the \emph{tag36h11} family, with a tag size of $6.4cm$ and an ID of 0.

\subsection{Results}
In the experiments, we considered two distinct scenarios. Scenario I involved commanding the rover to move forward at a constant velocity of $1\frac{cm}{s}$, while the manipulator remained stationary. Both cameras continuously captured synchronized AprilTag poses from the target object. In Scenario II, the rover was commanded to move at $0.5\frac{cm}{s}$, and the hand-mounted camera moved along with the target object. In this case, measurements from the two cameras were not always available or synchronized. In both cases, the process noise covariance matrices were set to $\vect{Q}_{n_v} = 10^{-2} \vect{I}_3$ and $\vect{Q}_{n_w} = 10^{-5} \vect{I}_3$ while the initial measurement noise covariance was $
\vect{R}_i \!=\!
\left[\begin{smallmatrix}
10^{-6} \vect{I}_3 & \vect{0}_3 \\
\vect{0}_3 & 10^{-3} \vect{I}_3
\end{smallmatrix}\right]
$. The discretization time interval was $\Delta t \!=\! 0.066s$ for Scenario I, corresponding to the sensor sampling rate, and $\Delta t \!=\! 0.25s$ for Scenario II. The initial covariance matrix in both scenarios was set to $\vect{P}(0|0) \!=\! 10^{-2} \vect{I}_{15}$ while the forgetting factors were $f_Q \!=\! f_R \!=\! 0.999$ for Scenario I, and $f_Q \!=\! 0.990$, $f_R \!=\! 0.950$ for Scenario II.

\begin{figure}[!h]
  \unitlength=0.5in
   \centering 
    \subfloat[]{
   \includegraphics[width=1.6in]{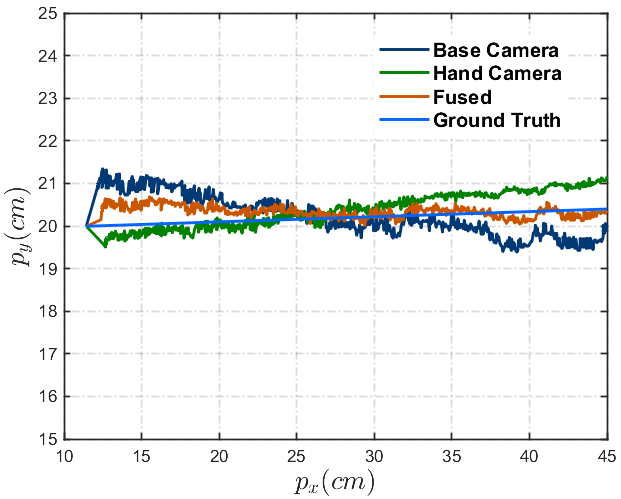}
  }~
      \subfloat[]{ \includegraphics[width=1.6in]{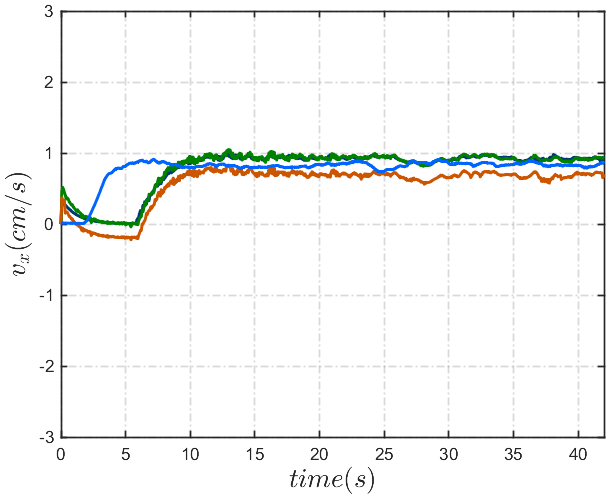}}
    \\\caption{Results of scenario I; (a) target trajectory, and (b)  its velocity.}\label{fig: results1}
\end{figure}

\begin{figure}[!h]
  \unitlength=0.5in
   \centering 
    \subfloat[]{
   \includegraphics[width=1.6in]{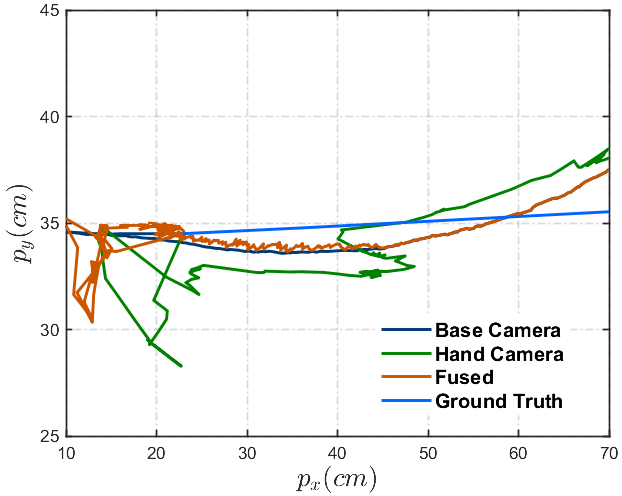}
  }~
      \subfloat[]{ \includegraphics[width=1.6in]{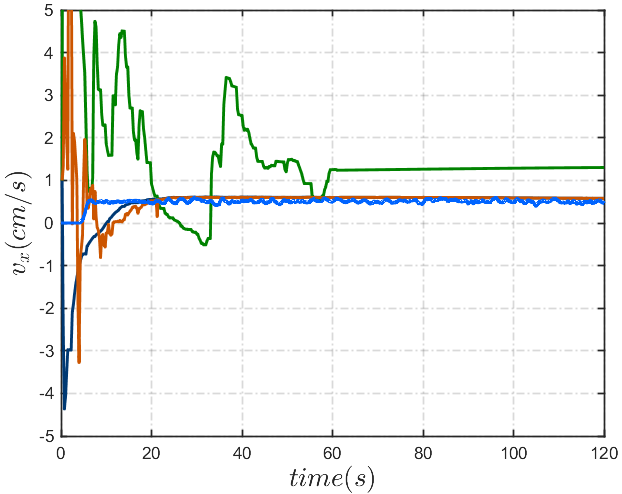}}
   \\
   \caption{Results of scenario II; (a) target trajectory, and (b)  its velocity.}\label{fig: results2}
\end{figure}

The position and velocity estimates are presented in Fig.~\ref{fig: results1} and Fig.~\ref{fig: results2}. In Scenario I, the fused states closely follow the ground truth, demonstrating the effectiveness of the fusion approach (Fig.~\ref{fig: results1}). This is primarily due to the consistency of both local filters throughout the estimation process, supported by frequent and synchronized measurements from the cameras. 

In contrast, Scenario II involves more challenging conditions (Fig.~\ref{fig: results2}). The base-mounted EKF is updated approximately $70\%$ of the time, while the hand-mounted EKF receives updates only $33\%$ of the time. Furthermore, the accelerations introduced by the arm’s movement have an inverse impact on AprilTag pose detection. To accommodate these uncertainties, the forgetting factors are set to lower values compared to Scenario I, allowing the filters to adapt more aggressively to the rapidly changing dynamics. The reduced availability of measurements, combined with increased non-Gaussian noise, leads to inconsistencies in the filters. These inconsistencies violate the underlying assumptions of the fusion algorithm, particularly during the initial stages of motion, which results in degraded performance. However, as the manipulator and rover transition to a more stable, near-constant velocity regime and more reliable data becomes available, the filter gradually converges and produces satisfactory results.

\begin{figure}[]
  \unitlength=0.5in
   \centering 
    \subfloat[]{
   \includegraphics[width=1.6in]{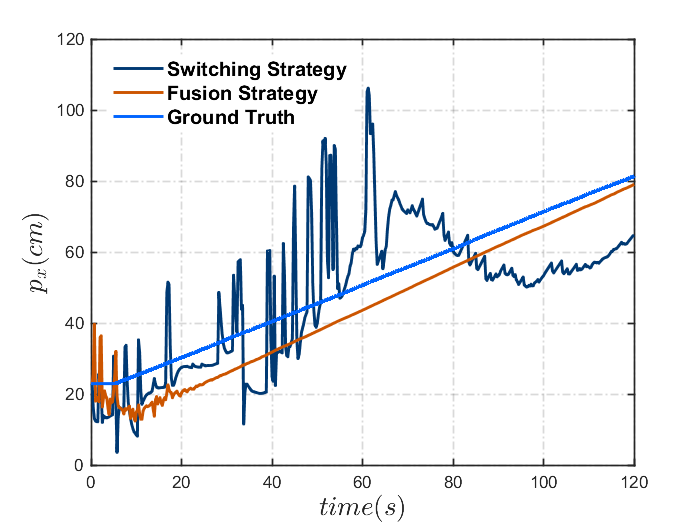}
  }~
      \subfloat[]{ \includegraphics[width=1.6in]{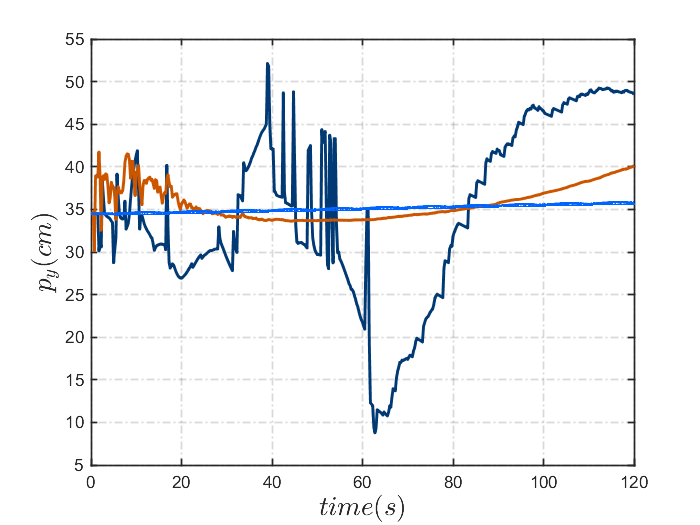}}\\
      ~
      \subfloat[]{ \includegraphics[width=1.6in]{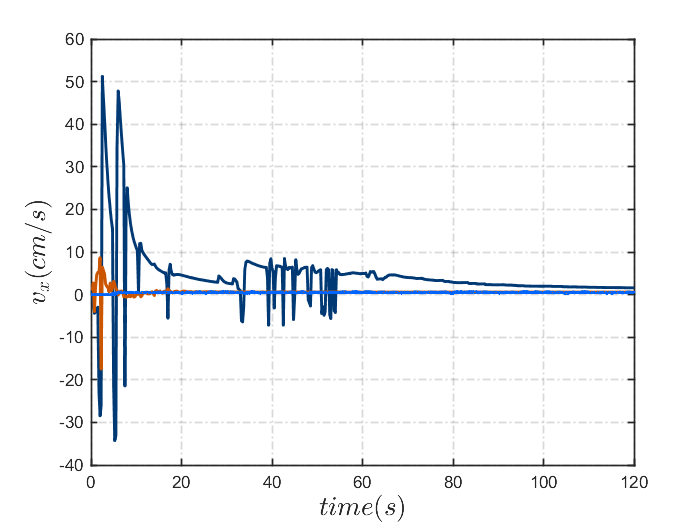}}~
      \subfloat[]{ \includegraphics[width=1.6in]{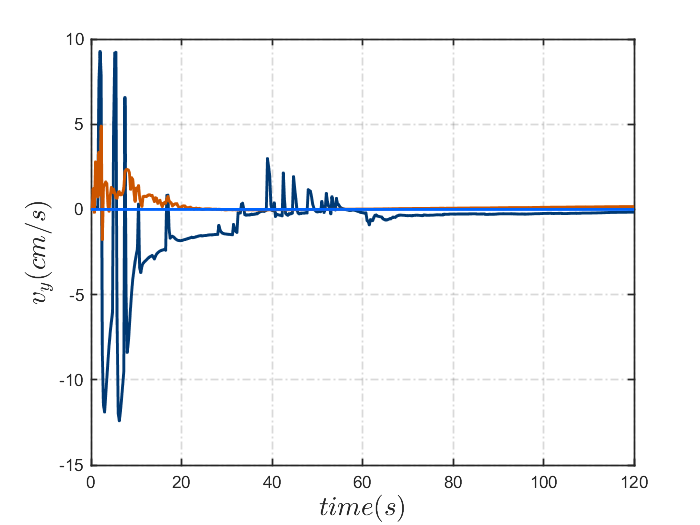}}
    \\\caption{Comparison between fusion and switching strategies in Scenario II; (a) fused forward position $p_x$, (b) fused lateral position $p_y$, (c) fused forward velocity $v_x$ and (d) fused lateral velocity $v_y$.}\label{fig:switching}
\end{figure}

To evaluate the effectiveness of the proposed fusion strategy, we compared it with a switching approach similar to~\cite{cuevas2018hybrid}, which alternates between pose measurements from the hand and base cameras. To this end, we implemented a switching filter for Scenario II, where the EKF updates its state only when a measurement (either from the hand or the base camera) is available. All parameters were kept identical for both estimation frameworks. The results are presented in Fig.~\ref{fig:switching}. The superior performance of the fusion strategy is attributed to its ability to combine estimates from the base and hand EKFs based on their respective confidence levels. In contrast, the switching approach updates solely based on the availability of measurements from the vision sensors. In the challenging conditions of Scenario II, this strategy leads to degraded estimation performance.

\begin{figure}[]
  \unitlength=0.5in
   \centering 
    \subfloat[]{
   \includegraphics[width=1.6in]{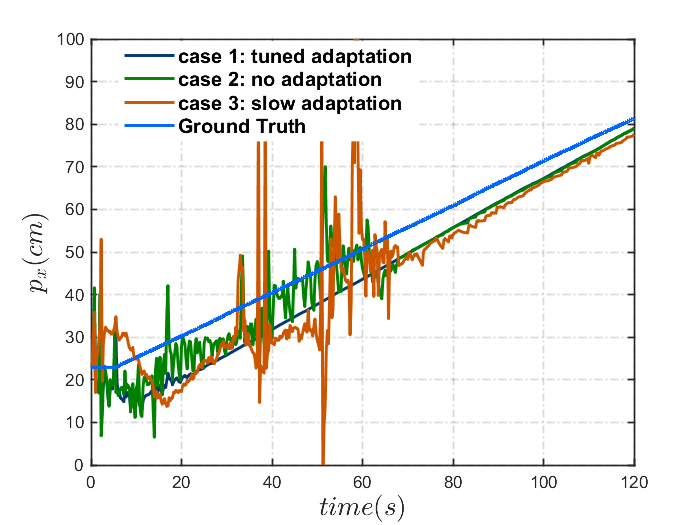}
  }~
      \subfloat[]{ \includegraphics[width=1.6in]{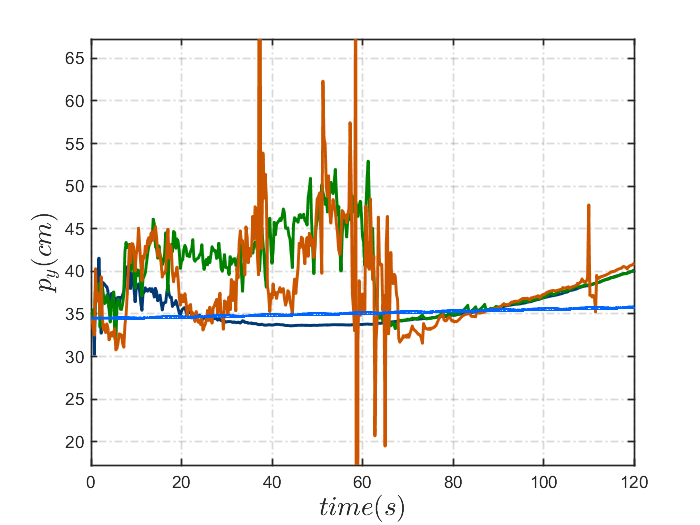}}\\\subfloat[]{
   \includegraphics[width=1.6in]{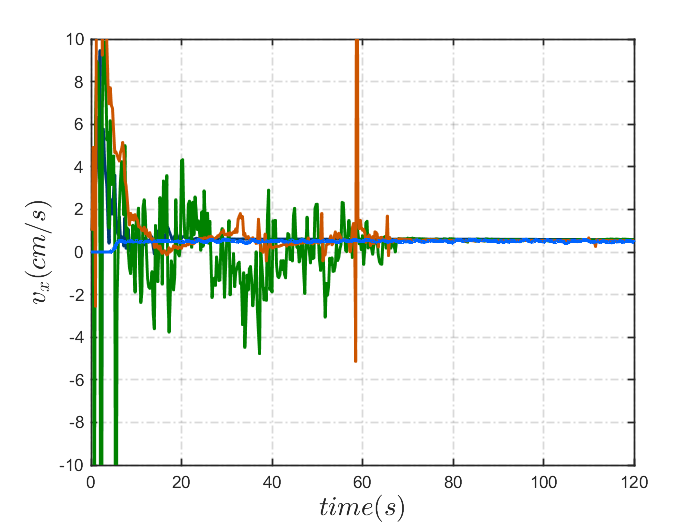}
  }~
      \subfloat[]{ \includegraphics[width=1.6in]{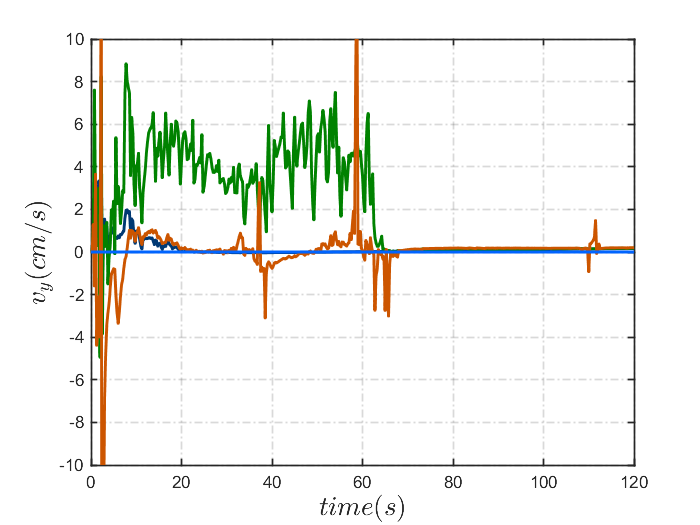}}\\
      \caption{Comparison study of the  adaptation factors in Scenario II for three cases; (a) fused forward position $p_x$, (b) fused lateral position $p_y$, (c) fused forward velocity $v_x$ and (d) fused lateral velocity $v_y$.}\label{fig:adapt}
\end{figure}

In our experiments, we evaluated the effect of adaptation factors on the fusion performance  through three cases (See Fig.~\ref{fig:adapt}). Case 1 uses tuned adaptation factors reported earlier and produces smooth and consistent state estimates. This case achieves a good balance between responsiveness to measurements and robustness against noise and model uncertainties. In Case 2, adaptation is disabled, resulting in noisier state estimates. Without adaptation, the filter cannot adequately inflate its covariance to account for unmodeled dynamics or uncertainties, leading to inconsistency. Case 3 applies a low adaptation factor of 0.8 for all filters, which leads to even noisier and highly unstable results. In this case, the filter overreacts to measurement noise, causing large fluctuations in the estimated states. These results highlight the importance of carefully tuning the adaptation factors to maintain filter consistency and ensure reliable fusion performance.

\section{Conclusions}\label{sec-con}
In this paper, we addressed the problem of pose and velocity estimation and fusion for a moving target using a dual-vision setup. The proposed method was developed within the Lie group framework and incorporated a correlation-aware, adaptive decentralized estimation scheme on matrix Lie groups. Experimental results demonstrated that when sufficient measurements are available and local filters remain consistent, the proposed fusion approach achieves high estimation accuracy. The framework was implemented on a UFactory xArm 850 manipulator and benchmarked against state-of-the-art methods for dual-view pose estimation. Results confirm the superiority of our approach, particularly under challenging conditions involving high accelerations and sudden jerks in manipulator motion. 

Since non-Gaussian noise, unmodeled dynamics, and unexpected uncertainty can lead to filter inconsistency, a key direction for future work is to explore data-driven estimation and fusion methods that relax Gaussian assumptions and shortcomings of model-based approaches. Furthermore, while this work relied on fiducial markers and known grasping references, future efforts will focus on estimating object pose and grasping frames directly from visual features using machine learning, informed by object geometry and dynamics. Finally, integrating complementary sensing modalities, such as proximity sensors, may enhance short-range estimation.


\bibliographystyle{IEEEtran}
\bibliography{References}

\begin{thebibliography}{10}
\providecommand{\url}[1]{#1}
\csname url@samestyle\endcsname
\providecommand{\newblock}{\relax}
\providecommand{\bibinfo}[2]{#2}
\providecommand{\BIBentrySTDinterwordspacing}{\spaceskip=0pt\relax}
\providecommand{\BIBentryALTinterwordstretchfactor}{4}
\providecommand{\BIBentryALTinterwordspacing}{\spaceskip=\fontdimen2\font plus
\BIBentryALTinterwordstretchfactor\fontdimen3\font minus
  \fontdimen4\font\relax}
\providecommand{\BIBforeignlanguage}[2]{{%
\expandafter\ifx\csname l@#1\endcsname\relax
\typeout{** WARNING: IEEEtran.bst: No hyphenation pattern has been}%
\typeout{** loaded for the language `#1'. Using the pattern for}%
\typeout{** the default language instead.}%
\else
\language=\csname l@#1\endcsname
\fi
#2}}
\providecommand{\BIBdecl}{\relax}
\BIBdecl

\bibitem{sanderson1980image}
A.~Sanderson, ``Image-based visual servo control using relational graph error
  signals,'' in \emph{Proc. IEEE Int. Conf. on Cybernetics and Society, 1980},
  1980, pp. 1074--1077.

\bibitem{hager1996tutorial}
G.~Hager, S.~Hutchinson, and P.~Corke, ``Tutorial tt3: A tutorial on visual
  servo control,'' in \emph{Held at ICRA 1994 Conference}, 1996.

\bibitem{janabi2010comparison}
F.~Janabi-Sharifi, L.~Deng, and W.~J. Wilson, ``Comparison of basic visual
  servoing methods,'' \emph{IEEE/ASME Transactions on Mechatronics}, vol.~16,
  no.~5, pp. 967--983, 2010.

\bibitem{labbe2020cosypose}
Y.~Labb{\'e}, J.~Carpentier, M.~Aubry, and J.~Sivic, ``Cosypose: Consistent
  multi-view multi-object 6d pose estimation,'' in \emph{European Conference on
  Computer Vision}.\hskip 1em plus 0.5em minus 0.4em\relax Springer, 2020, pp.
  574--591.

\bibitem{wen2020se}
B.~Wen, C.~Mitash, B.~Ren, and K.~E. Bekris, ``se (3)-tracknet: Data-driven 6d
  pose tracking by calibrating image residuals in synthetic domains,'' in
  \emph{2020 IEEE/RSJ International Conference on Intelligent Robots and
  Systems (IROS)}.\hskip 1em plus 0.5em minus 0.4em\relax IEEE, 2020, pp.
  10\,367--10\,373.

\bibitem{janabi1998automatic}
F.~Janabi-Sharifi and W.~J. Wilson, ``Automatic grasp planning for visual-servo
  controlled robotic manipulators,'' \emph{IEEE Transactions on Systems, Man,
  and Cybernetics, Part B (Cybernetics)}, vol.~28, no.~5, pp. 693--711, 1998.

\bibitem{janabi2010kalman}
F.~Janabi-Sharifi and M.~Marey, ``A kalman-filter-based method for pose
  estimation in visual servoing,'' \emph{IEEE transactions on Robotics},
  vol.~26, no.~5, pp. 939--947, 2010.

\bibitem{liang2024adaptive}
X.~Liang, ``Adaptive position-based visual servoing of robot manipulators,'' in
  \emph{International Conference on Intelligent Robotics and
  Applications}.\hskip 1em plus 0.5em minus 0.4em\relax Springer, 2024, pp.
  341--356.

\bibitem{he2022deep}
Y.~He, J.~Gao, and Y.~Chen, ``Deep learning-based pose prediction for visual
  servoing of robotic manipulators using image similarity,''
  \emph{Neurocomputing}, vol. 491, pp. 343--352, 2022.

\bibitem{ribeiro2023second}
E.~G. Ribeiro, R.~Q. Mendes, M.~H. Terra, and V.~Grassi, ``Second-order
  position-based visual servoing of a robot manipulator,'' \emph{IEEE Robotics
  and Automation Letters}, vol.~9, no.~1, pp. 207--214, 2023.

\bibitem{bauml2010kinematically}
B.~B{\"a}uml, T.~Wimb{\"o}ck, and G.~Hirzinger, ``Kinematically optimal
  catching a flying ball with a hand-arm-system,'' in \emph{2010 IEEE/RSJ
  International Conference on Intelligent Robots and Systems}.\hskip 1em plus
  0.5em minus 0.4em\relax IEEE, 2010, pp. 2592--2599.

\bibitem{daravina2024visual}
G.~C. Daravi{\~n}a, J.~L. Valencia, G.~A. Holguin, H.~F. Quintero, E.~A. Ariza,
  and D.~Vergara, ``Visual servoing and kalman filter applied to parallel
  manipulator 3-rrr,'' \emph{Electronics}, vol.~13, no.~14, p. 2703, 2024.

\bibitem{maniatis2017best}
C.~Maniatis, M.~Saval-Calvo, R.~Tylecek, and R.~B. Fisher, ``Best viewpoint
  tracking for camera mounted on robotic arm with dynamic obstacles,'' in
  \emph{2017 International Conference on 3D Vision (3DV)}.\hskip 1em plus 0.5em
  minus 0.4em\relax IEEE, 2017, pp. 107--115.

\bibitem{flandin2000eye}
G.~Flandin, F.~Chaumette, and E.~Marchand, ``Eye-in-hand/eye-to-hand
  cooperation for visual servoing,'' in \emph{Proceedings 2000 ICRA. Millennium
  Conference. IEEE International Conference on Robotics and Automation.
  Symposia Proceedings (Cat. No. 00CH37065)}, vol.~3.\hskip 1em plus 0.5em
  minus 0.4em\relax IEEE, 2000, pp. 2741--2746.

\bibitem{kermorgant2011multi}
O.~Kermorgant and F.~Chaumette, ``Multi-sensor data fusion in sensor-based
  control: Application to multi-camera visual servoing,'' in \emph{2011 IEEE
  International Conference on Robotics and Automation}.\hskip 1em plus 0.5em
  minus 0.4em\relax IEEE, 2011, pp. 4518--4523.

\bibitem{dune2007one}
C.~Dune, E.~Marchand, and C.~Leroux, ``One click focus with
  eye-in-hand/eye-to-hand cooperation,'' in \emph{Proceedings 2007 IEEE
  International Conference on Robotics and Automation}.\hskip 1em plus 0.5em
  minus 0.4em\relax IEEE, 2007, pp. 2471--2476.

\bibitem{assa2014virtual}
A.~Assa and F.~Janabi-Sharifi, ``Virtual visual servoing for multicamera pose
  estimation,'' \emph{IEEE/ASME Transactions on Mechatronics}, vol.~20, no.~2,
  pp. 789--798, 2014.

\bibitem{lippiello2007position}
V.~Lippiello, B.~Siciliano, and L.~Villani, ``Position-based visual servoing in
  industrial multirobot cells using a hybrid camera configuration,'' \emph{IEEE
  Transactions on Robotics}, vol.~23, no.~1, pp. 73--86, 2007.

\bibitem{cuevas2018hybrid}
H.~Cuevas-Velasquez, N.~Li, R.~Tylecek, M.~Saval-Calvo, and R.~B. Fisher,
  ``Hybrid multi-camera visual servoing to moving target,'' in \emph{2018
  IEEE/RSJ International Conference on Intelligent Robots and Systems
  (IROS)}.\hskip 1em plus 0.5em minus 0.4em\relax IEEE, 2018, pp. 1132--1137.

\bibitem{zarei2024consistent}
M.~Zarei and R.~Chhabra, ``Consistent fusion of correlated pose estimates on
  matrix lie groups,'' \emph{IEEE Robotics and Automation Letters}, vol.~9,
  no.~7, pp. 6584--6591, 2024.

\bibitem{bourmaud2013discrete}
G.~Bourmaud, R.~M{\'e}gret, A.~Giremus, and Y.~Berthoumieu, ``Discrete extended
  kalman filter on lie groups,'' in \emph{21st European Signal Processing
  Conference}.\hskip 1em plus 0.5em minus 0.4em\relax IEEE, 2013, pp. 1--5.

\bibitem{chirikjian2011stochastic}
G.~S. Chirikjian, \emph{Stochastic models, information theory, and Lie groups,
  volume 2: Analytic methods and modern applications}.\hskip 1em plus 0.5em
  minus 0.4em\relax Springer Science \& Business Media, 2011, vol.~2.

\bibitem{murray2017mathematical}
R.~M. Murray, Z.~Li, and S.~S. Sastry, \emph{A mathematical introduction to
  robotic manipulation}.\hskip 1em plus 0.5em minus 0.4em\relax CRC press,
  2017.

\bibitem{barfoot2017state}
T.~D. Barfoot, \emph{State estimation for robotics}.\hskip 1em plus 0.5em minus
  0.4em\relax Cambridge University Press, 2017.

\bibitem{mehra1970adaptive}
R.~K. Mehra, ``On the identification of variances and adaptive kalman
  filtering,'' \emph{IEEE Transactions on Automatic Control}, vol.~15, no.~2,
  pp. 175--184, 1970.

\bibitem{mohamed1999adaptive}
A.~Mohamed and K.~Schwarz, ``Adaptive kalman filtering for ins/gps,''
  \emph{Journal of geodesy}, vol.~73, no.~4, pp. 193--203, 1999.

\bibitem{realSenseCam2025}
\BIBentryALTinterwordspacing
``Realsense depth camera,'' accessed: 2025-09-08. [Online]. Available:
  \url{https://www.intel.com/content/www/us/en/products/sku/128255/intel-realsense-depth-camera-d435/specifications.html}
\BIBentrySTDinterwordspacing

\bibitem{CamCalib2025}
\BIBentryALTinterwordspacing
``Camera calibration,'' accessed: 2025-09-08. [Online]. Available:
  \url{https://wiki.ros.org/camera_calibration}
\BIBentrySTDinterwordspacing

\bibitem{limo2025}
\BIBentryALTinterwordspacing
``Agilex limo robot,'' accessed: 2025-09-08. [Online]. Available:
  \url{https://global.agilex.ai/products/limo-pro}
\BIBentrySTDinterwordspacing

\bibitem{apriltag2025}
\BIBentryALTinterwordspacing
``Apriltag ros package,'' accessed: 2025-09-08. [Online]. Available:
  \url{https://wiki.ros.org/apriltag_ros}
\BIBentrySTDinterwordspacing

\end{thebibliography}

\end{document}